\documentclass[10pt, a4paper, twocolumn]{article} %

\usepackage[english]{babel} %

\usepackage{microtype} %

\usepackage{amsmath,amsfonts,amsthm} %

\usepackage[svgnames]{xcolor} %

\usepackage[small, labelfont={color=DarkGoldenrod,bf}, up, textfont=it]{caption} %

\usepackage{booktabs} %
\usepackage{float}
\usepackage{lastpage} %

\usepackage{graphicx} %
\usepackage{tikz}
\usetikzlibrary{backgrounds}
\usepackage{enumitem} %

\newsavebox{\picbox}

\setlist{noitemsep} %
\usepackage[framemethod=tikz]{mdframed}

\usepackage{geometry} %

\usepackage[T1]{fontenc} %
\usepackage[utf8]{inputenc} %

\usepackage{XCharter} %

\usepackage{fancyhdr} %
\fancypagestyle{firstpage}{ %
	\fancyhf{}
}

\newcommand{\authorstyle}[1]{{\large\usefont{OT1}{phv}{b}{n}#1}} %

\newcommand{\institution}[1]{{\footnotesize\usefont{OT1}{phv}{m}{sl}\color{Black}#1}} %

\usepackage{titling} %

\newcommand{\HorRule}{\color{DarkGoldenrod}\rule{\linewidth}{1pt}} %

\newcommand{\HorRuleRed}{\color{DarkRed}\rule{\linewidth}{1pt}} %

\pretitle{
	\vspace{-30pt} %
	\HorRule\vspace{10pt} %
	\fontsize{26}{32}\usefont{OT1}{phv}{b}{n}\selectfont %
	\color{DarkRed} %
}

\posttitle{\par\vskip 15pt} %

\preauthor{} %

\postauthor{ %
	\vspace{5pt} %
	\par\HorRule %
	\vspace*{-27pt} %
}

\usepackage{aurical}

\usepackage{lettrine} %
\usepackage{fix-cm}	%

\newcommand{\initial}[1]{ %
	\lettrine[lines=5,findent=16pt,nindent=0pt]{%
		\color{DarkRed}%
		{#1}%
	}{}%
}

\usepackage{xstring} %

\mdfdefinestyle{RedFrame}{%
    outerlinewidth=2pt,
    roundcorner=5pt,
    innertopmargin=\baselineskip,
    innerbottommargin=\baselineskip,
    innerrightmargin=10pt,
    innerleftmargin=10pt,
    linecolor=DarkRed,
    leftmargin=-0.4cm,
    rightmargin=-0.4cm,
    }

\usepackage[backend=biber,style=ieee,natbib=true,doi=false,isbn=false,url=false,eprint=false]{biblatex} %

\usepackage[autostyle=true]{csquotes} %

\usepackage{wrapfig}

\usepackage{bbding}

\usepackage{hyperref}
\usepackage{pdfpages}
\usepackage{eso-pic}

\graphicspath{{figs/}}

\title{\protect\centering A RoboStack Tutorial\\[0.2cm]\normalsize Using the Robot Operating System Alongside the Conda and Jupyter Data Science Ecosystems}

\author{
    {
       \protect\centering \protect\includegraphics[width=0.5\linewidth]{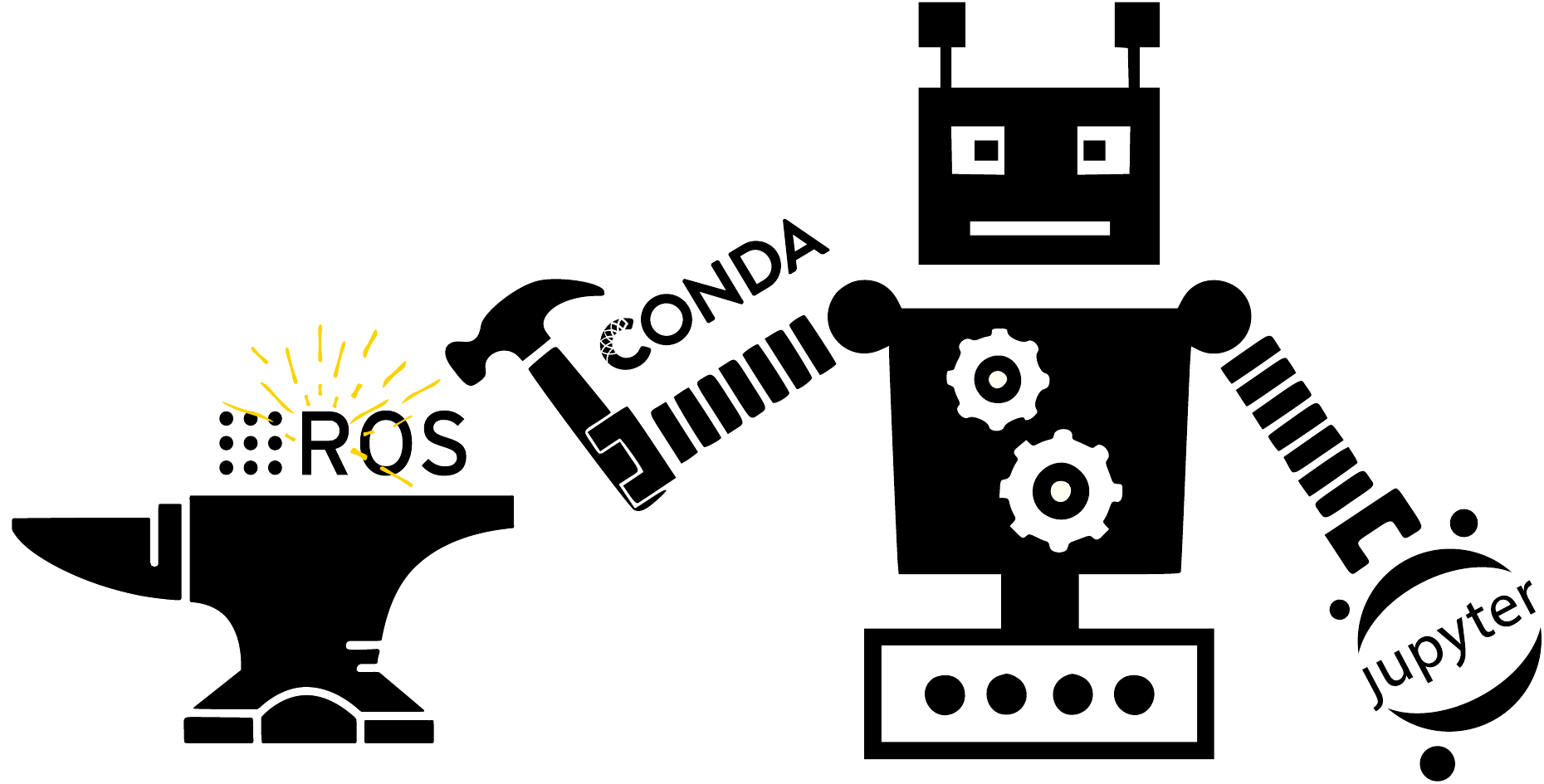}\\
       \authorstyle{By Tobias Fischer\textsuperscript{1}, Wolf Vollprecht\textsuperscript{2}, Silvio Traversaro\textsuperscript{3},\\Sean Yen\textsuperscript{4}, Carlos Herrero\textsuperscript{2} and Michael Milford\textsuperscript{1}}\\[0.5cm]
    }
	\textsuperscript{1}\institution{Queensland University of Technology, QUT Centre for Robotics, Brisbane, Australia}\\
	\textsuperscript{2}\institution{QuantStack, Berlin, Germany}\\
	\textsuperscript{3}\institution{Fondazione Istituto Italiano Di Tecnologia (Italian Institute of Technology), Genova, Italy}\\
	\textsuperscript{4}\institution{Microsoft, Redmond, Washington, USA}
}

\date{}

\begin{document}

\AddToShipoutPicture*{%
     \AtTextUpperLeft{%
         \put(-3,11){
           \begin{minipage}{\textwidth}
              \textit{Preprint version\\Final version available at }\url{https://doi.org/10.1109/MRA.2021.3128367}
           \end{minipage}}%
     }%
}

\maketitle %

\thispagestyle{fancy}

\initial{T}he Robot Operating System (ROS) has become the de-facto standard middleware in the robotics community~\cite{ROS}. ROS bundles everything, from low-level drivers to tools that transform between coordinate systems, to state-of-the-art perception and control algorithms. One of ROS's key merits is the rich ecosystem of standardized tools to build and distribute ROS-based software. 

In a parallel development, a large suite of scientific libraries in the machine learning, deep learning, and data science domains has emerged: NumPy, SciPy, Pandas, OpenCV, the Natural Language Toolkit, PyTorch, and TensorFlow to name a few. These libraries have wide ranging applications including computer vision, computer graphics and natural language processing -- all of which are also highly relevant in robotics.

However, interfacing ROS with these scientific libraries has traditionally been a complex endeavor. The official support of ROS is limited to very few platforms and operating systems that often come with ``outdated'' software packages, which raises the issue that running ROS and the latest scientific libraries -- which in turn frequently rely on ``bleeding edge'' dependencies -- is often impractical, especially if multiple of those libraries should run at the same time. The same observations hold when wanting to run ROS alongside other robotics libraries, such as the new Ignition simulation tools~\cite{ferigo2020gym} and the Robotics Toolbox for Python~\cite{roboticstoolboxpython}.

Taking these versioning issues aside, another opportunity that has seen limited attention thus far is a tight integration of ROS with Jupyter (with notable first steps being introduced in~\cite{cervera2019roslab}). Jupyter is a web-based interactive development platform that offers complex widgets, including video streams and sliders, a flexible and configurable user interface to support various workflows, as well as plotting tools; all in an environment that joins code, interface, and documentation using \emph{Notebooks}. Using ROS alongside Jupyter will be particularly useful in cloud robotics applications, both for development and monitoring purposes.

\begin{table*}[t]
\caption{Summary table and glossary}
\vspace*{-0.2cm}
\resizebox{0.99\linewidth}{!}{
\begin{tabular}{ll}
Term         & Description              \\\midrule
Boa & Fast build tool for Conda packages\\
Conda & Cross-platform, language-agnostic package manager and environment manager\\
Conda channel & Software repository where packages are stored \\
Conda recipe & Files describing package dependencies and build instructions to create a Conda package\\
Conda-forge channel & Community-led Conda channel containing over 10,000 packages with recent versions\\
RoboStack channel & Channel build on top of Conda-forge that contains ROS packages\\
Jupyter & Web-based interactive development platform\\
Jupyter Notebook & Open document format that can contain code, equations, visualizations and narrative text\\
Jupyter-ROS & A collection of Jupyter interactive widgets inspired by Qt and RViz\\
JupyterLab & Next-generation user interface within the Jupyter ecosystem\\
JupyterLab-ROS & A set of ROS plugins for JupyterLab\\
Mamba & Fast drop-in replacement for the conda tool\\
Mambaforge & An installer for Conda that uses the Conda-forge channel by default and contains Mamba\\
Vinca & Packaging tool similar to bloom that creates Conda recipes from ROS package.xml files
\end{tabular}
}
\label{tab:glossary}
\end{table*}

{
\noindent\HorRuleRed
\\[0.2cm]
\color{DarkRed}
\textbf{%
In summary, RoboStack provides Conda packages for ROS, and a variety of ROS related plugins for Jupyter notebooks. Table~\ref{tab:glossary} contains a summary of the terminology and proposed tools.
}
\\
\HorRuleRed
}

This tutorial introduces \emph{RoboStack}, addressing both opportunities, i.e.~installing machine learning and other scientific libraries alongside ROS on multiple platforms with ease (RoboStack targets \emph{any} Linux distribution, Windows and macOS), and integration of ROS with Jupyter. Part 1 of this tutorial shows the reader how to install -- and, for developers, how to build -- ROS packages using the Conda package manager. Conda not only provides a wide range of scientific libraries, but also comes with a rich ecosystem that is of interest to the robotics community, including \emph{conda lock} to create reproducible environments and \emph{conda constructor} to provide easy-to-use installers. Part 2 of this tutorial introduces the reader to a collection of JupyterLab extensions for ROS, including plugins for live plotting, debugging, and robot control, as well as tight integration with Zethus, an RViz like visualization tool.

\begin{table}[t]
\caption{Comparison of different package managers}
\resizebox{0.99\linewidth}{!}{
\begin{tabular}{lccc}
\multicolumn{1}{c}{Property}         & apt & pip  & conda              \\\midrule
Select precise versions               & $\times$     & o\protect\footnotemark & \checkmark \\
Binary libraries (C++/C...) & \checkmark & $\times$                                           & \checkmark \\
Bleeding edge software versions       & $\times$     & \checkmark                   & $\checkmark$ \\
Decentralized packages                & $\times$     & \checkmark                   & $\checkmark$
\end{tabular}
}
\label{tab:packagemanagers}
\end{table}

\section*{\color{DarkGoldenrod}Part 1: ROS+Conda}
\begin{figure}[t]
  \centering
  \includegraphics[width=0.99\linewidth]{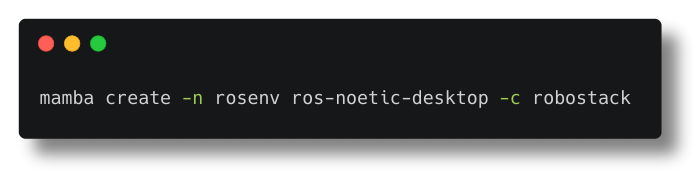}
  \caption{Creating a new environment named rosenv with ros-noetic-desktop installed. This is even easier than installing ROS using apt on Ubuntu (the recommended means of installing ROS), as adding the custom RoboStack channel is done by simply appending `-c robostack`, as compared to the traditional installation process which requires setting up a sources.list and needing to obtain keys for this server. It also works in the same way for all supported platforms, i.e.~Linux, macOS and Windows. RoboStack recommends using Mambaforge to install Conda, as this sets Conda-forge as the default channel. If this is not the case, one also has to append `-c conda-forge'. It is recommended to install additional packages as detailed in \url{https://robostack.github.io/GettingStarted.html}}%
  \label{fig:ROS_install}%
\end{figure}

Before this tutorial delves into the details of the installation of ROS via the Conda package manager (which in the simplest case could as simple as executing a single command, see Figure~\ref{fig:ROS_install}), we introduce common terminology in the ``Conda land'' (Table~\ref{tab:glossary} provides a summary) and list some of the many advantages of this means of installation. For readers that want to get started in a flash, we note that the main instructions are summarized in the figures contained within this tutorial.
\footnotetext{A dependency resolver was recently added to pip but still has some flaws.}

\subsection*{Conda terminology}
Conda is a cross-platform, language-agnostic package manager, that enables building packages with ease for all common operating systems, i.e.~Linux, Windows and macOS, as well as various architectures, i.e.~x86, ARM and PowerPC -- including cross-compilation. Furthermore, and contrary to the popular pip package manager for Python packages (and its associated PyPI software repository), Conda manages binary dependencies and has mechanisms that ensure compatibility across packages so that a large number of packages can be installed side-by-side without conflicts. A comparison of apt, pip and conda can be found in Table~\ref{tab:packagemanagers}.

While Conda refers to the package manager itself, Conda \emph{channels} refer to locations (or software repositories) where packages are stored. One of the most popular channels is the community-led Conda-forge~\cite{conda_forge_community_2015_4774217}, which contains thousands of scientific packages written in Python, C++, Julia and more, including deep learning frameworks like PyTorch and TensorFlow, and computer vision libraries like OpenCV. Thanks to generous continuous integration services offered by Microsoft Azure and other providers, these packages are automatically built in the cloud for all the above mentioned architectures. Conda-forge is decentralized: each package is maintained and build within its own repository. Opposed to PyPI, it is common that a package is maintained by a third-party, i.e.~the maintainer does not have to be the source-code author. Another popular channel is Bioconda, which provides over 3,000 Conda software packages that are concerned with bioinformatics~\cite{gruning2018bioconda}.
{
\noindent\HorRuleRed
\\[0.2cm]
\color{DarkRed}
\textbf{%
Conda-forge is one of the fastest-growing and most successful open-source projects of the past years. It was started in 2015 and already hosts over 10,000 packages, has more than 1,600 individual contributors and roughly 200 million package downloads per month.
}
\\
\HorRuleRed
}

\subsection*{Conda adaptation in robotics}
While Conda and Conda-forge are widely used within the data science, machine learning, deep learning and computer vision communities, they have not yet seen a significant uptake in the robotics community. In the case of ROS, this is largely due to conflicts and incompatibilities in the internal mechanisms of ROS and Conda. Generally one of the best practices when using Conda is to install as many dependencies as possible using Conda, which is not easily feasible in the case of ROS. While wrapper scripts could work around those conflicts by re-setting environment variables\footnote{Indeed, such workarounds have been created: \url{https://github.com/rickstaa/.ros_conda_wrapper}}, such workarounds are brittle and are not well suited in scenarios where robustness is required. Furthermore, using these workarounds only work on Linux, while there is a growing demand to run ROS on Windows and macOS, too.

\subsection*{RoboStack and its advantages}
\begin{wrapfigure}{R}{3.5cm}
{
\HorRuleRed
\\[0.2cm]
\color{DarkRed}
\textbf{%
RoboStack combines the best of the data-science and robotics worlds to help researchers and developers building custom solutions for their academic and industrial projects.
}
\\
\HorRuleRed
}
\end{wrapfigure}

To circumvent these issues, this tutorial introduces \emph{RoboStack} that tightly couples ROS with Conda and Jupyter, a web-based interactive computational environment that affords scientific computing. The new RoboStack Conda channel provides ROS packages that are built with dependencies from Conda-forge, so that ROS can be installed alongside the latest data-science and machine-learning packages with ease. 

The authors of the tutorial have further contributed and maintain a wide range of other robotics libraries on Conda-forge, including the Ignition Libraries and Gazebo for robot simulation, Ogre as 3D graphics engine, the Flexible Collision Library (FCL), the Bullet physics engine, and the Dynamic Animation and Robotics Toolkit (DART).

As Conda is also an environment manager that can isolate environments, multiple ROS versions (currently Melodic\footnote{\label{melodicnote}The RoboStack effort is concentrated on ROS Noetic as this has a much better support for Python 3 compared to ROS Melodic. Thus ROS Melodic is very experimental at this stage. Similarly, for ROS2 we concentrate our efforts on ROS2 Galactic with experimental support for Foxy.}, Noetic and ROS2 Foxy\textsuperscript{\ref{melodicnote}} and Galactic) can run simultaneously on one machine. The integration of ROS with Conda has further benefits: RoboStack provides pre-compiled binaries for Linux, Windows and macOS, as well as the ARM architecture for e.g.~the Raspberry Pi. Note that RoboStack's ROS binaries are decoupled from specific distributions -- while the ``vanilla'' ROS Noetic only runs on Ubuntu 20.04, the Conda binaries run on any Linux distribution with glibc 2.12 or newer (e.g.~Fedora, Ubuntu 12.04 and newer, CentOS, Linux Mint). ROS can now be installed using an easy-to-use installer that does not require any dependencies (not even Conda). If Conda is already installed, creating a new environment named \emph{rosenv} with ROS installed is as simple as running a single command, as shown in Figure~\ref{fig:ROS_install}.

\noindent In essence, having ROS on Conda provides several \emph{advantages}, namely updated dependencies, platform-agnostic installation, no need for root permissions, environment isolation, use of binaries, reproducible environments, and easy package creation.

\begin{description}[noitemsep,topsep=0pt]
    \item[Updated dependencies:] Conda-forge typically has cutting-edge versions of packages. This enables using recent versions of dependencies on older distributions such as Ubuntu 16.04. There is no need anymore to wait for Linux distributions to ``catch up''.
    \item[Platform-agnostic installation:] RoboStack provides a unified way to install ROS binaries on Windows, macOS or any Linux distribution (Fedora, Ubuntu 16.04, Ubuntu 19.10, ...). This also simplifies support, as all dependencies are installed via Conda rather than the distribution-specific package manager. In fact, the conda-constructor tool provides a means of a ``one-click installer'' that installs Conda and ROS packages: All the user has to do is choosing an installation directory; no prior installation of Conda is required.
    \item[No need for root permissions:] All software installed by Conda is installed and used in an arbitrary directory such as the user's home directory. This enables installation of ROS, other robotics packages and their dependencies on a system in which one does not have root access (such as a shared workstation or a high-performance computer). As a technical detail, Conda makes packages relocatable: prefix information is only inserted into the package at installation time.
    \item[Environment isolation:] Conda provides a way to seamlessly install different ROS distributions side-by-side -- for example one could have two environments, one with ROS Melodic, and the other with ROS Noetic. One can even run ROS packages from both environments at the same time using different terminal sessions.
    \item[Use of binaries:] Conda distributes its packages as binaries. This has the advantage that installing heavy dependencies such as OpenCV, PCL, Qt and Gazebo on Windows takes just a few minutes, as opposed to hours when using vcpkg where packages are compiled locally. This is also useful when producing Docker images: installing dependencies via Conda takes a few seconds, which reduces the time necessary to regenerate Docker images. At this point we also note that Conda binaries run ``bare metal'', i.e.~there is no performance overhead.
    \item[Reproducible environments:] Conda has built-in support for installing exactly the same version of packages on the same or other machines, up to the patch version. This is a significant feature for reproducibility in scientific research, which has seen increasing interest in recent years~\cite{bonsignorio2017new,perkel2020challenge,cervera2018try}. Conda also enables upgradable environments, where certain packages are locked to specific versions, while others can be updated to more recent versions. 
    \item[Easy package creation:] It is much simpler to create Conda packages than it is to create Debian packages. It is also simple to set up custom (or even private) channels to distribute them.
\end{description}

\subsection*{RoboStack weaknesses}
\noindent With respect to the classical way of installing ROS using the apt package manager, RoboStack also provides some \emph{disadvantages} and \emph{weaknesses}. There are three main points in this regard. 

Firstly, RoboStack is incompatible with C++ or Python libraries that can only be installed via apt. While the underlying Conda-forge hosts over 10,000 packages, this is still a long way off the approximately 50,000 packages in Debian and Ubuntu. 

Secondly, the RoboStack project does not contain all available ROS packages. Thus far, the RoboStack project contains around 1250 out of the 1500 released packages on ROS Noetic for the Linux platform. Most of the remaining packages either require further dependencies to be ported to Conda-forge (see first weakness), are abandoned, or do not yet work with Python 3. Most other platforms (macOS, Windows, and Linux for ARM) have builds for desktop-full (which includes the communication protocols, the ``general robotics'' libraries of ROS, perception libraries, simulators and visualization and debugging tools), the navigation stack and a range of other packages as detailed in \url{https://robostack.github.io/noetic.html} -- macOS currently has builds for 1000 packages, and there are 700 RoboStack packages for Windows. Most recently, builds of RViz (ROS's visualization tool) and its dependencies have been added for the new ``Apple Silicon'' architecture, with many more to be added in the next months. 

The third weakness is the possibility of security issues. Conda-forge and ROS are community-driven, which means that anyone can contribute packages. Therefore, a malicious actor could potentially introduce malware into one of these packages, which would then be re-distributed in RoboStack. In addition, malware could be introduced when distributing packages to the RoboStack channel. Mitigations are in place at all levels to minimize the risk of such security vulnerabilities: Conda-forge and ROS have security sub-teams (in the case of Conda-forge, at least one member of this sub-team is a maintainer of the most frequently used packages), Conda-forge performs security scans of produced artifacts, and all modifications that are introduced to RoboStack packages are publicly visible in our repositories and reviewed by the core RoboStack team. The necessary rights to add and modify packages in the RoboStack channel is also only granted to the RoboStack core team.

\begin{figure}[t]
  \centering
  \includegraphics[width=0.99\linewidth]{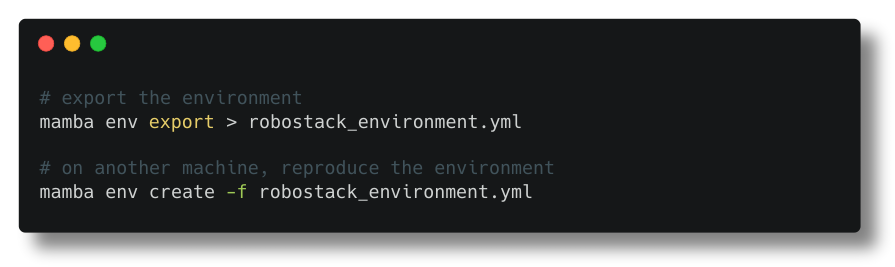}
  \includegraphics[width=0.99\linewidth]{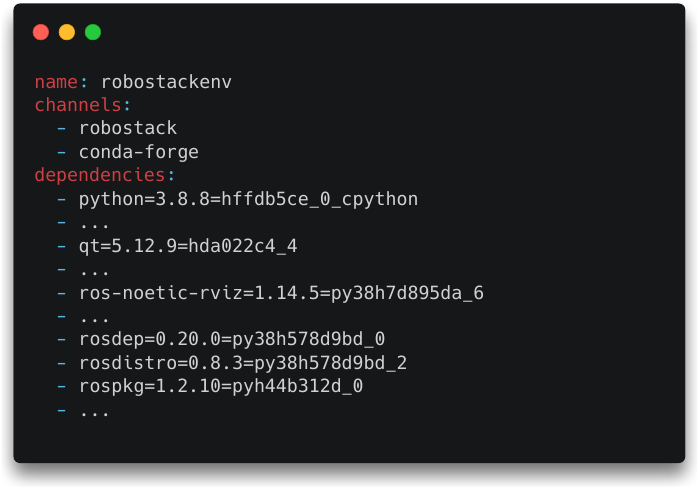}
  \caption{Reproducing an environment using Conda (top) and created robostack\_environment.yaml file (bottom).}%
  \label{fig:reproducing_simple}%
\end{figure}

\subsection*{Reproducibility with Conda / RoboStack}
We would now like to detail how environments can be reproduced with Conda. Firstly, reproducing the exact same environment using Conda is effortless. One can simply export the environment to a file with `conda env export', and then read in this file to obtain a copy of the environment with the exact same versions and builds of all packages, as shown in Figure~\ref{fig:reproducing_simple}. While this allows for excellent reproducibility, it has the drawback that upgrading individual packages becomes near-impossible. On the other hand, one could drop the version and build number in the dependencies list of the environment.yaml file, to simply list `python' instead of `python=3.8.8=hffdb5ce\_0\_cpython' and so forth. While this would work well to obtain the latest packages in the future, doing so is inadequate for reproducibility, as new versions might not be compatible with the version for which tests were performed.

There are several improvements that aim for reproducible \emph{and} upgradable environments. A first step is to use so-called versioned dependencies, i.e.~specifying `python=3.8' but not pinning to a specific build. Then upgrading a subset of the packages is easy (e.g.~changing `python=3.8' to `python=3.9' while leaving the other package versions untouched), but it comes with the risk that dependencies of dependencies have changed in an incompatible way. The conda-lock utility combines the best of both worlds. It takes a file with versioned dependencies as input (created using `conda env export -{}-no-builds`), and outputs a locked dependency file. It has the added benefit that the locked dependency files can be created for multiple platforms, so that a Linux environment can be reproduced -- almost exactly -- on macOS and Windows. Such environments facilitate benchmarking tools for robotics.

\subsection*{Building ROS packages using RoboStack}
The appendix of this tutorial details several key advances of interest for developers that were made so that ROS packages can be build on top of the Conda-forge ecosystem. It describes \emph{Vinca}, a packaging tool that creates Conda recipe from ROS package.xml files, \emph{Mamba}, a fully compatible fast drop-in replacement for Conda, and \emph{Boa}, a fast package builder to build Conda packages. It also contains instructions and examples on how to add new packages to RoboStack, and how to use RoboStack in continuous integration for existing ROS packages to easily build and test packages on multiple platforms. Note that we use \emph{Mamba} in the remainder of this tutorial as it significantly speeds up the creation of Conda environments and installation of new packages.

\section*{\color{DarkGoldenrod}Part 2: ROS+Jupyter}
Historically, the ROS (Robot Operating System) community has relied on Qt for building complex user interfaces. Nowadays, the Jupyter Notebook, JupyterLab and the ipywidgets framework offer a compelling alternative. As in Part 1 of this tutorial, we first introduce terminology and list the advantages of using ROS with Jupyter, followed by detailed installation and usage instructions.

\subsection*{Jupyter terminology}
Jupyter is a web-based interactive development platform. A key component to the Jupyter stack is the Notebook, an open-source web application that allows creating and sharing documents that contain live code, equations, visualizations and narrative text. JupyterLab is the next-generation user interface for Jupyter notebook, integrating Jupyter notebooks with text editors, terminals and custom extensions. JupyterLab-ROS is a set for ROS plugins for JupyterLab, enabling for example to control the ROS core within JupyterLab, so that development, visualizations and monitoring all come together in a single interface. Complex widgets using browser technology are possible using ipywidgets: from JavaScript sliders to 3D with WebGL to real time video streaming with WebRTC. Thanks to the ipywidgetification, these visualizations can be created without writing any JavaScript (which makes it significantly easier to use than e.g.\ YARP-JS~\cite{ciliberto2017connecting}).

\subsection*{Advantages of using Jupyter and ROS}
A robotics development environment using Jupyter  essentially runs in a web browser, which provides a high degree of flexibility. Indeed, Jupyter Notebooks are already frequently used in tasks that are related to robotics, ranging from simple ones like data cleaning and transformation, to complex ones like numerical simulation, statistical modeling, data visualization, and machine learning. 

As it can be tedious to have different platforms for interacting with robots and processing the data obtained from these, this second part introduces Jupyter-ROS and JupyterLab-ROS that enable an integrated workflow. We note that these packages are made available via the RoboStack Conda channel, and thus the same benefits that were listed in the Part 1 of this article also apply to this second part. Some of the additional advantages include:

\begin{enumerate}
    \item Any web browser can be used -- Jupyter is not bound to Linux, and no Qt applications need to be compiled.
    \item Applications do not need to run locally. They can run on a remote server, without any manual setup or installation procedure.
    \item All of these tools are plugin-based. This enables diverse use cases, from simple interactive scripting to visualization of large datasets. In fact, in JupyterLab everything is a plugin, including the core components to the application. Anyone can produce a ``remix'' of core and third-party JupyterLab extensions to their needs.
    \item In the spirit of ROS, one of the keys to Jupyter's success is that the project was built upon well-documented and specified protocols and file formats that anyone could implement.
    \item Jupyter is a multi-stakeholder project, not backed by a single corporation, but by a community of developers at a variety of companies, universities, as well as individual contributors. Such multi-stakeholder projects can offer a high degree of generalization and robustness because of the input from a diverse group of contributors.
\end{enumerate}

With regards to the reproducibility of Jupyter Notebooks, they have range of advantages, which makes them increasingly popular in many scientific and technical fields~\cite{beg2021using}. However, they also have a key weakness, namely the arbitrary execution order of the code cells~\cite{wang2020assessing}, considering that the user wanting to replicate the notebook might have to guess the authors' order of the cell executions.

\subsection*{ROS \& Jupyter: Target domains}
\begin{wrapfigure}{R}{3.5cm}
{
\HorRuleRed
\\[0.2cm]
\color{DarkRed}
\textbf{%
RoboStack's plugins turn JupyterLab into a \emph{cloud robotics command station}.
}
\\
\HorRuleRed
}
\end{wrapfigure}

One of the aims of Jupyter-ROS and JupyterLab-ROS is to be the go-to platform for interdisciplinary projects, in particular those that integrate machine learning or computer vision with robotics. Machine learning and computer vision researchers that are interested in applying their models to robots now have the possibility of connecting a robot to a development environment that they are familiar with. 

Another opportunity for robotics researchers and educators alike is in using Notebooks to create and share experiments in a dynamic and interactive development environment that allows quick prototyping and exploratory analysis.

We also advocate for the use of Jupyter-ROS and JupyterLab-ROS widgets, as well as the Jupyter ecosystem as a whole, in cloud robotics~\cite{wan2016cloud,kehoe2015survey, ray2016internet}. In such a cloud setting, some of the software powering one or multiple robots will run on high-end computers in data centers. Jupyter and JupyterLab can then be used for both development and monitoring purposes. The entire process of robot initialization, parameter tuning and remote supervision of one or multiple robots can be done from the flexible and powerful JupyterLab interface. Robot customers will be able to login to a single user-friendly interface, without having to install any custom software on their machine, or run a specialized operating system.

\begin{figure}[t]
  \centering
  \includegraphics[width=0.99\linewidth]{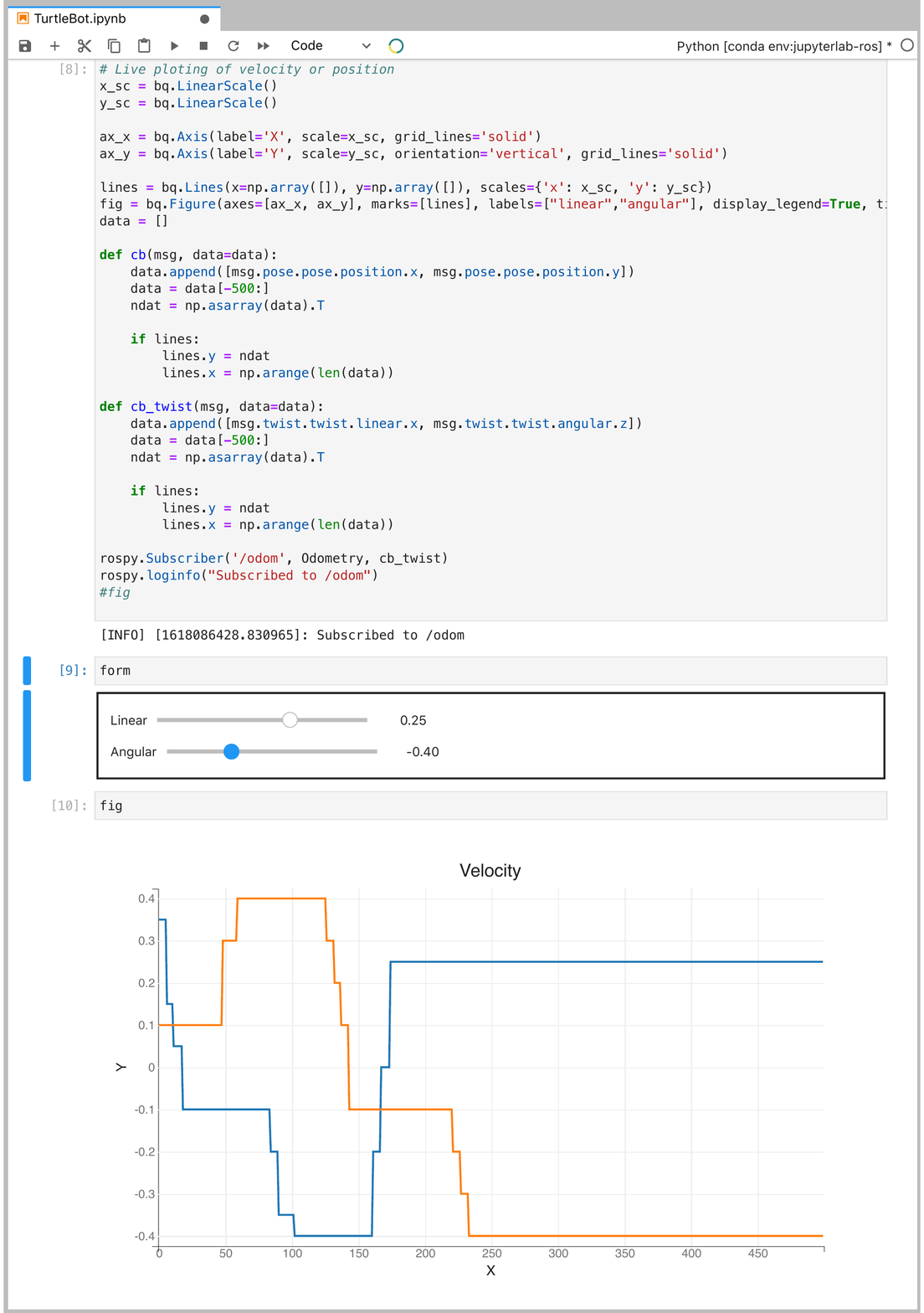}
  \caption{This example shows a simple widget to control the linear and rotational velocity of a mobile robot.}%
  \label{fig:robotcontrol}%
\end{figure}

\subsection*{ROS \& Jupyter: Challenges}
By default ROS does not play nicely with Jupyter. For example, rospy (the Python client library for ROS) is inherently multi-threaded: when executing a cell that creates a new subscriber, every new ROS topic subscriber spawns a new thread -- a thread that keeps running in the background even after the execution of the cell ends. By default, if this thread contains debug printouts, and another cell is run, the printouts of the thread will be printed to the new cell, since the thread is still running in the background.

Another problem is that a Jupyter Notebook is a nice console that can display rich and structured outputs, but everything that a user runs stays in memory until the kernel is restarted. Imagine writing ROS programs in the Python interactive console. When re-executing a cell, the previous state of the cell is not overwritten unless the variable is overwritten. Similarly, a thread running in the background does not stop unless we stop the thread or reset the kernel. Therefore, if a cell that creates a new ROS topic subscriber is executed twice, there will be two background threads; but the variable that used to point to the first thread now points to the second thread. That means the first subscriber cannot be accessed anymore (and thus not stopped) unless the kernel is reset.

\subsection*{Jupyter-ROS plugins}
\begin{figure}[t]
\centering
\includegraphics[width=0.99\linewidth]{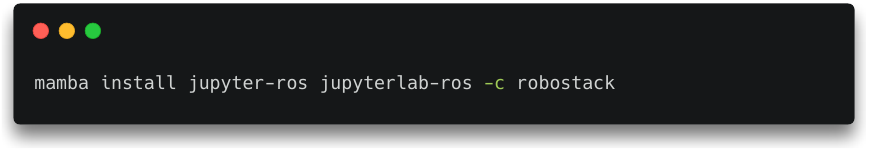}
\caption{Installation of Jupyter-ROS and JupyterLab-ROS}%
  \label{fig:installjupy}%
\end{figure}

To overcome these problems, the RoboStack authors created Jupyter-ROS and JupyterLab-ROS. Jupyter-ROS is a suite of plugins to the Jupyter ecosystem to make working with ROS inside Jupyter a breeze. While Jupyter-ROS is applicable across the whole Jupyter ecosystem, JupyterLab-ROS is a set of widgets specifically designed for JupyterLab, the next-generation user interface within the Jupyter ecosystem. Both can be easily installed as shown in Figure~\ref{fig:installjupy}.

The previously mentioned problem of flooding user interfaces due to debug printouts is solved as follows: when subscribing to a topic using Jupyter-ROS, the function returns an ipywidget with a start/stop button and a dedicated output area for debug prints. Internally this re-routes all printouts from the subscriber thread to this Jupyter cell, and provides full control over the thread (by being able to stop and restart it at any time). 

\begin{figure}[t]
  \centering
  \includegraphics[width=0.99\linewidth]{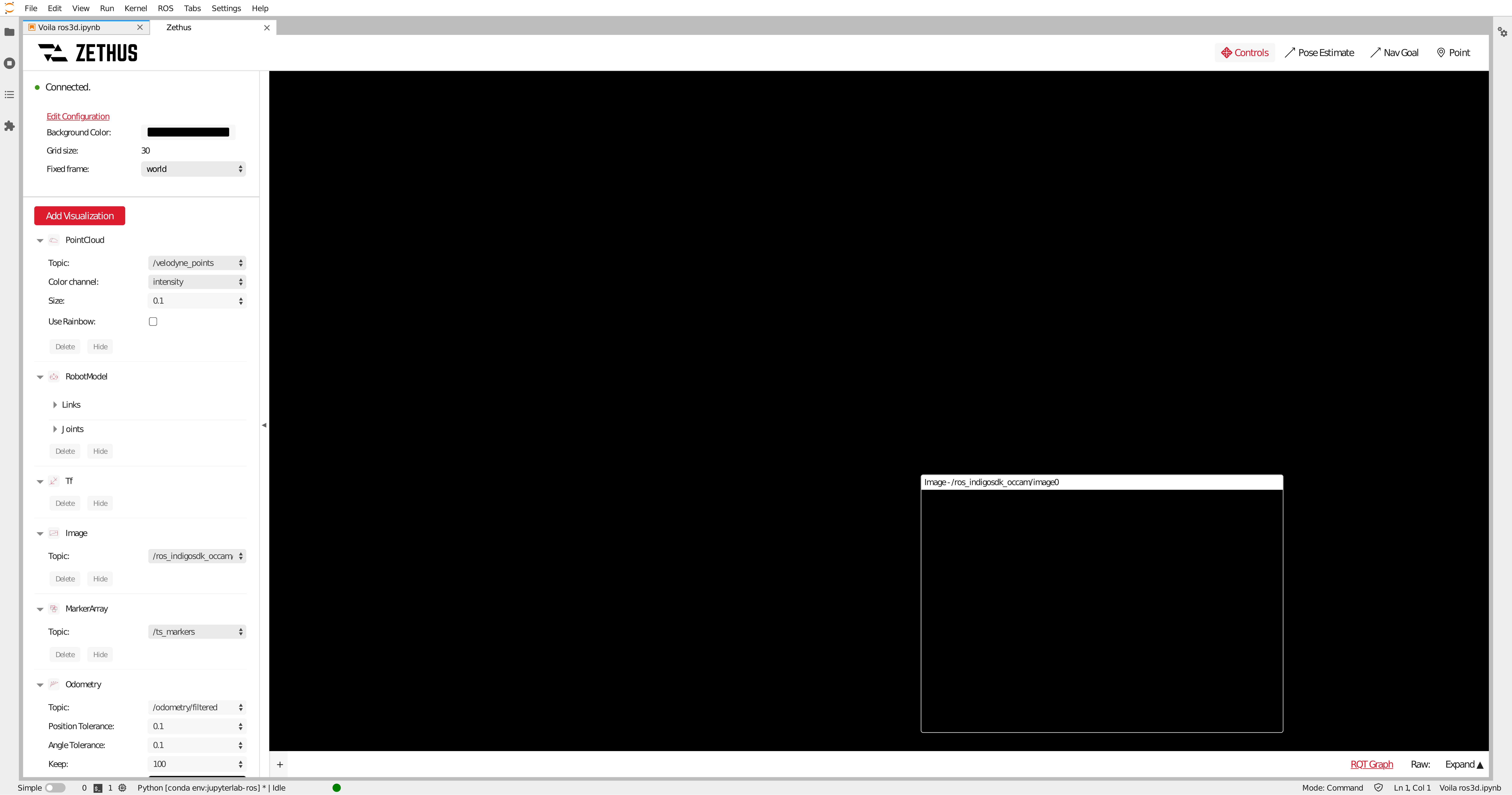}
  \caption{The plugin for Zethus provides an RViz like visualization tool within JupyterLab.}%
  \label{fig:zethus}%
\end{figure}

\begin{figure}[t]
  \centering
  \includegraphics[width=0.99\linewidth]{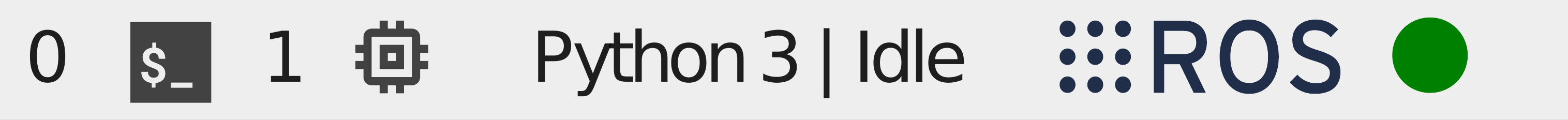}
  \caption{JupyterLab-ROS status bar.}%
  \label{fig:statusbar}%
\end{figure}

Jupyter-ROS further leverages ipywidgets and bqplot to show real time plots from ROS messages in Notebooks. Specifically, bqplot is a ``jupyter-native'' solution to plotting, with functionalities similar to what rqt\_plot offers in ROS. Jupyter-ROS allows users to select a subset of fields from a ROS message that will then be automatically plotted in a bqplot. Jupyter-ROS can also be used to obtain flexible, intuitive user-controlled inputs using forms. Figure~\ref{fig:robotcontrol} shows an example to control a simulated TurtleBot, and plot its velocity over time, all in a Jupyter Notebook.

While bqplot is well suited for 2D plots, oftentimes ROS users want to visualize the robot's environment in 3D. The go-to tool in the ROS ecosystem is RViz, a very powerful 3D visualization tool. Some of RViz's functionality has already been ported over to the web browser as part of the RobotWebTools effort~\cite{toris2015robot}. We extended the impressive work of the RobotWebTools to provide 3D ROS Jupyter widgets. They can currently provide RViz like visualizations for different data types, such as laser scans, robot trajectories, and 3D (URDF) models of the robot. 

\subsection*{JupyterLab-ROS plugins \& debuggging}
All previously described extensions are part of Jupyter-ROS and work in any Jupyter Notebook. We now introduce extensions that are tailored for JupyterLab, the next-generation interface within Jupyter. While the 3D ROS Jupyter visualizations are directly shown within the Notebook, Zethus provides a separate widget that supports nearly all the display types that RViz supports, as shown in Figure~\ref{fig:zethus}. Zethus also provides an info panel that displays the raw messages in realtime, and a web-based version of rqt\_graph for visualizing the ROS node graph. Zethus was originally developed by Rapyuta Robotics~\cite{mohanarajah2014rapyuta} and then adapted to work within JupyterLab.

JupyterLab-ROS makes the setup of WebSocket connections completely transparent to users by automatically starting a rosmaster and rosbridge\_server. A status bar widget (see Figure~\ref{fig:statusbar}) allows to start and stop the rosmaster server with a simple click and displays its state in real-time\footnote{An existing ROS master can be used by exporting the `ROS\_MASTER\_URI' before launching JupyterLab.}. Internally this widget runs a launch file which can be changed in the JupyterLab settings, for example to launch additional nodes. Another JupyterLab-ROS widget handles rosbag files to easily re-play rosbag files or record new rosbags.

\begin{figure}[t]
  \centering
  \includegraphics[width=0.99\linewidth]{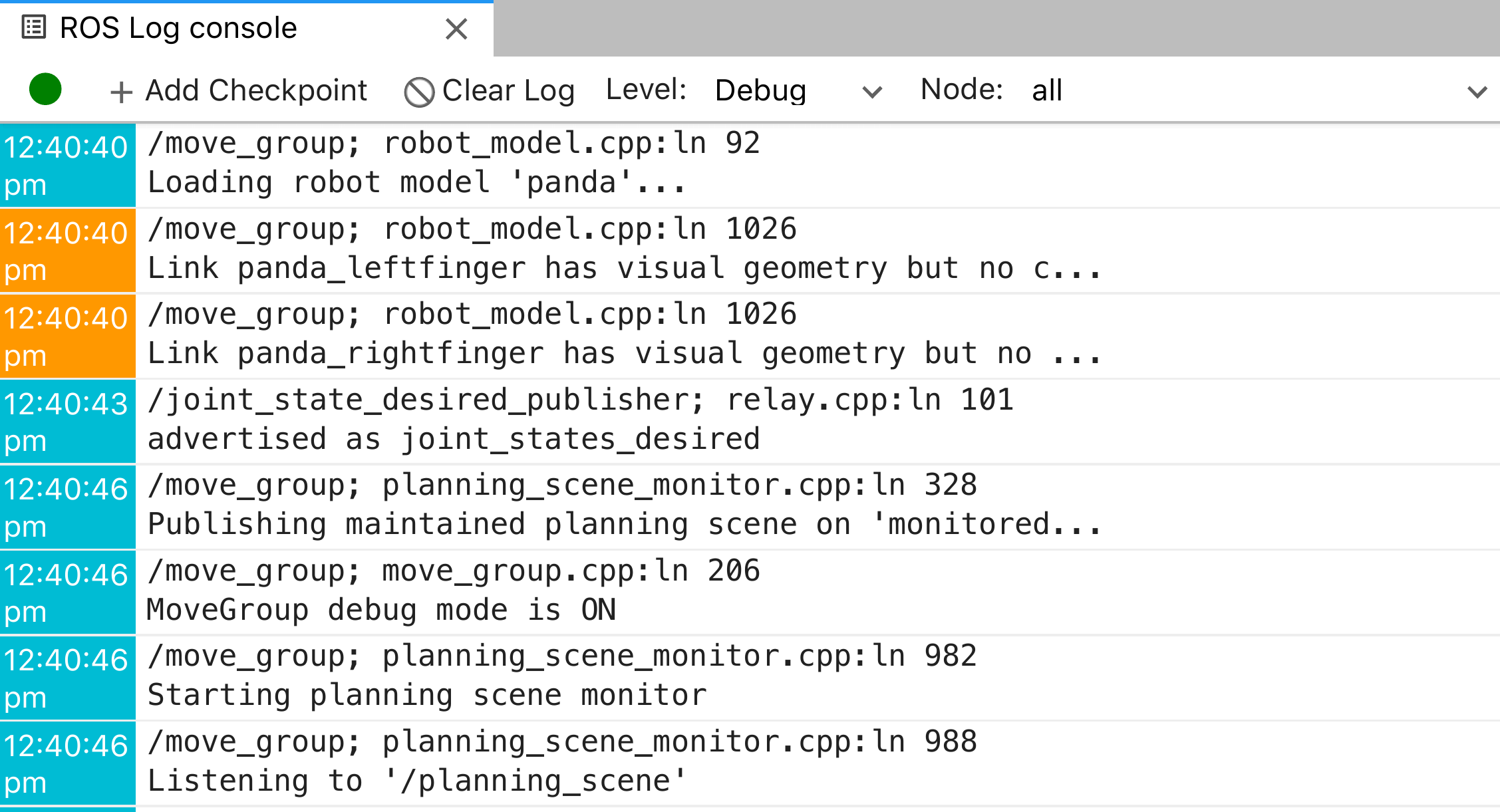}
  \caption{Logging in JupyterLab. One can specify the debug level, the node of interest, clear the log and highlight times by adding checkpoints.}%
  \label{fig:logconsole}%
\end{figure}

A robotics station would be incomplete without a tool for debugging. For that purpose, ROS provides rosconsole, a package which allows developers to send messages to the rosout topic so that they are available on every node. A console window gives access to these debugging messages in JupyterLab and provides additional features such as filtering by level and by node, as shown in Figure~\ref{fig:logconsole}.

\subsection*{Voilà: Converting Notebooks to web-apps}
The last package that this tutorial introduces is Voilà. In the basic sense, Voilà takes a Jupyter Notebook and removes all source code so that only the widgets (i.e.~input forms and visualizations) remain. Voilà can be used within a ROS launch file to display a specialized web-app for a robot environment (similar to RViz). Contrary to RViz, this web-app will be accessible from any computer that has a web-browser installed (such as a remote Windows computer). Another use case of Voilà is to present visualizations to a non-technical audience.

Voilà creates a standalone web application with a Python backend, whereby the backend consists of callbacks from ROS or from the widgets. As with the widgets in the Jupyter Notebook, the widgets in Voilá automatically sync the state between the frontend and backend. 

Note that Voilà is applicable beyond ROS and has not been designed for RoboStack in particular. Voilà is indeed a good example of existing functionality -- in this case within the Jupyter ecosystem -- that can now be used by ROS developers because of the tight coupling of ROS with Jupyter.

\begin{figure}[H]
\centering
\includegraphics[width=0.99\linewidth]{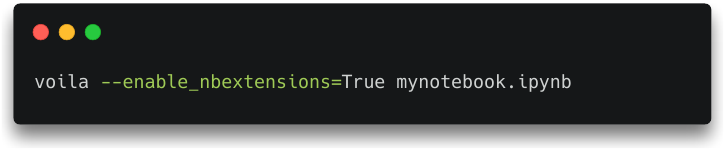}
 \caption{Running a Jupyter Notebook as a standalone Voilà web application.}%
  \label{fig:voila_run}%
\end{figure}

To run a Notebook as a Voilà application, one can either use the ``Open with Voilà in a new browser tab'' shortcut within JupyterLab, or execute a single command as shown in Figure~\ref{fig:voila_run}.

\begin{figure}[H]
  \centering
  \includegraphics[width=0.99\linewidth]{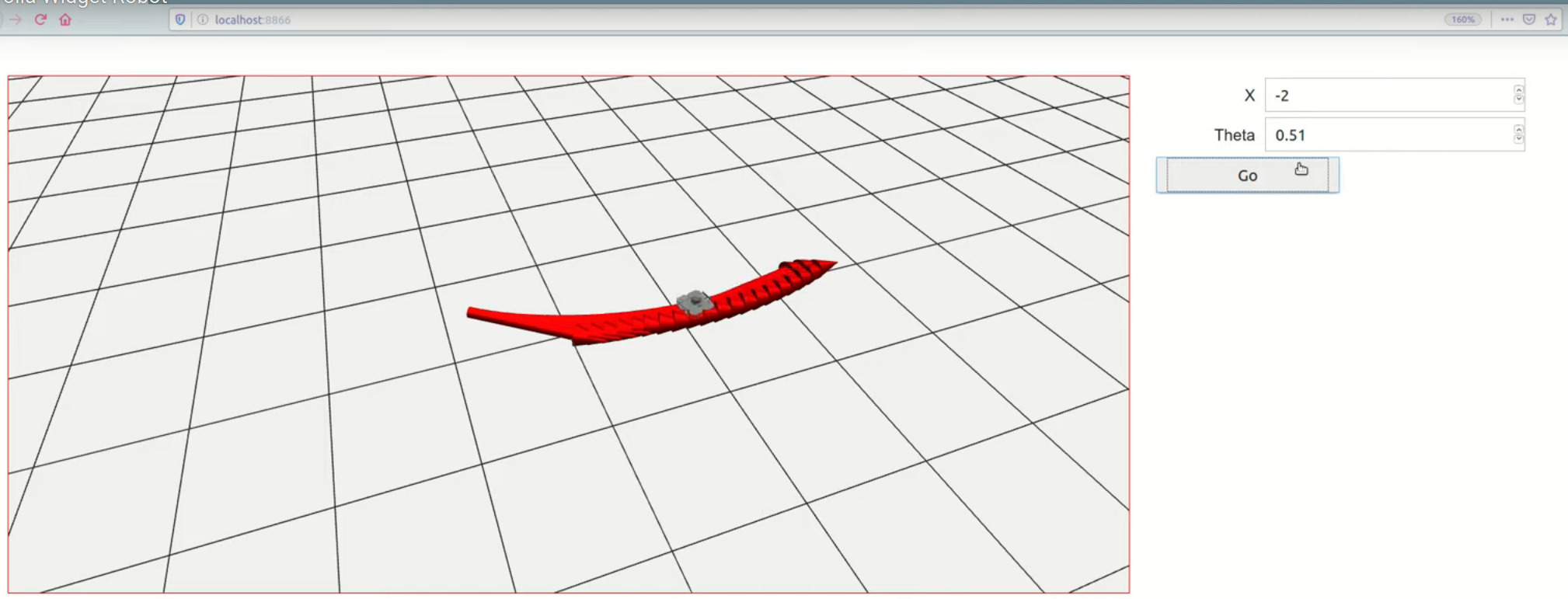}
  \caption{Example of a Voilà application in a web browser.}%
  \label{fig:voila}%
\end{figure}

After Voilà finishes executing all of the Notebook's cells, the browser tab contains a view of the widgets (just like it would in the Notebook). All buttons and callbacks will work just like they do in the Notebook, but no source code is displayed. An example Voilà application is shown in Figure~\ref{fig:voila}.

\section*{\color{DarkGoldenrod}RoboStack GitHub Organization}
RoboStack is currently being developed through a collaboration between academics at the Queensland University of Technology and the Italian Institute of Technology, and industrial partners at QuantStack and Microsoft. All code is freely and publicly available in a designated GitHub organization: \url{https://github.com/RoboStack}, in addition to the binary Conda packages which are available in the RoboStack channel: \url{https://anaconda.org/RoboStack}. The repositories contain guidelines for contribution. RoboStack aims for a friendly and inclusive community, and follows the NumFocus code of conduct.

\section*{\color{DarkGoldenrod}Acknowledgments}
The authors would like to thank all contributors to the RoboStack project. 

This work received funding from the Australian Government, via grant AUSMURIB000001 associated with ONR MURI grant N00014-19-1-2571. Tobias Fischer and Michael Milford acknowledge continued support from the Queensland University of Technology (QUT) through the Centre for Robotics.

The paragraph on reproducible and upgradable Conda environments was informed by Itamar Turner-Trauring's blog post on \url{https://pythonspeed.com/articles/conda-dependency-management/}. 

The cover logo has been designed using resources from Flaticon.com (authored by Freepik), conda-forge.com (the anvil is licensed under CC BY 4.0), and ROS.org (the ROS logo). Conda and the Conda logo are trademarked by Anaconda Inc. The Project Jupyter logo is Copyright by the Project Jupyter Contributors.

\addtolength{\textheight}{-13cm}
\addtolength{\footskip}{13cm}
\printbibliography[title={\color{DarkGoldenrod}References}]

\end{document}


\maketitle 

\thispagestyle{fancy}

The appendix of this RoboStack tutorial is aimed towards developers that are concerned with building ROS packages using Conda, which brings unique challenges. The key challenge with packaging ROS is that rather than building a single package with a handful of dependencies (which is the case for most packages on Conda-forge), ROS has thousands of packages with hundreds of dependencies -- some of them being external dependencies (such as PCL and Bullet) and others being dependencies on ROS packages. This is not only challenging because of potentially long build times if packages are build one after the other, but also because of Application Binary Interface (ABI) compatibility. The RoboStack project made several key advances around package management to overcome these challenges, all of which are applicable beyond the use case for ROS. We detail these advances, along with other information of interest for developers, in the following.


\renewcommand{\LettrineFontHook}{\Fontskrivan}

\initialsmall{1}{1}{0}
\textbf{The first key advance is the Vinca tool which automatically generates Conda recipes for ROS packages.} ROS packages have the advantage that dependencies are listed within package.xml files. Typically the rosdistro tool handles the conversion from dependency names within the package.xml files to package names that can be e.g.~installed using the apt system package manager on Ubuntu. For RoboStack, a similar mapping file was created that maps dependency names to package names in Conda-forge. Vinca then combines the latest package information obtained by the rosdistro Python API with this mapping file and a list of packages that should be build to generate Conda recipes. As some packages need Conda-specific changes, Vinca also allows the use of patches that can modify the source code before building the package. In the future, we aim for tight integration with the bloom release automation tool that is used in the ROS ecosystem to create Ubuntu packages.

\begin{figure*}[t]
  \centering
  \includegraphics[width=0.99\linewidth]{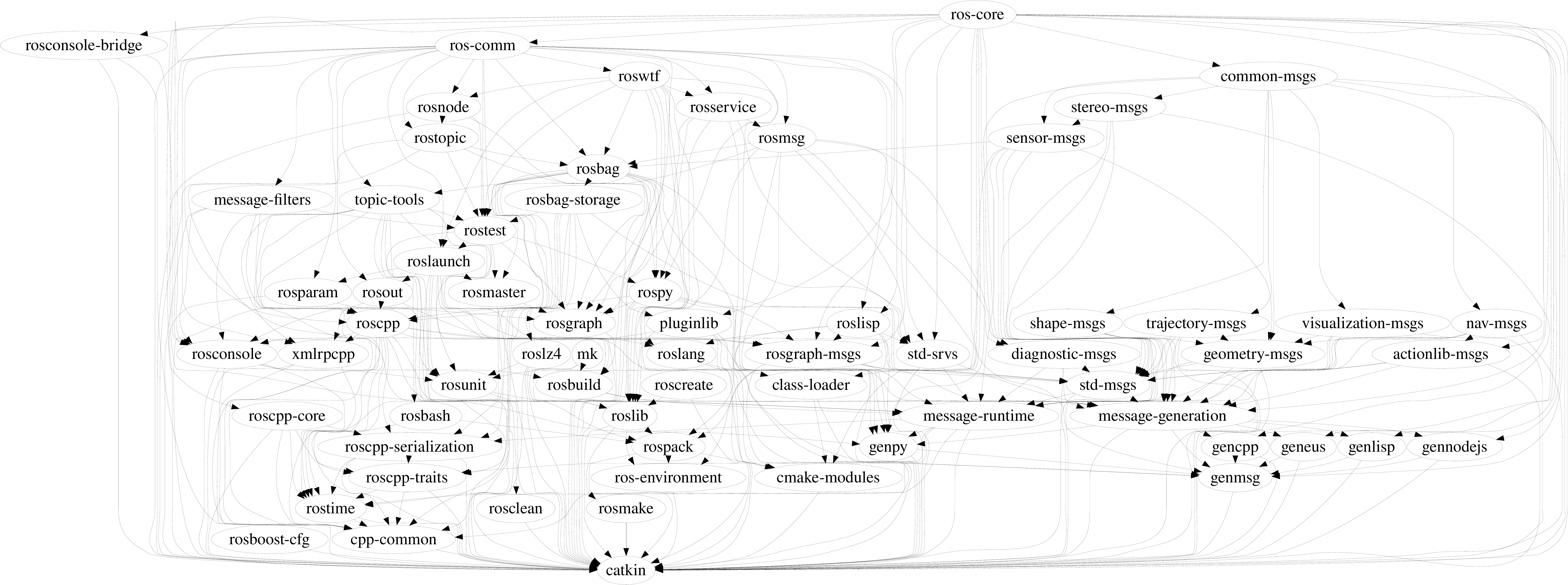}
  \caption{Dependency graph up to the ros-core meta-package. Each node represents a package, and each edge represents a dependency upon another package. ros-core depends on over 175 packages, whereby around 100 packages are pulled in from Conda-forge (such as Boost and Numpy) and the remaining packages are dependencies on other ROS packages such as rosbag and rostopic.}%
  \label{fig:dependencygraph}%
\end{figure*}

\initialsmall{2}{0.9}{0}
\textbf{The second key advance is the introduction of parallelism in the build pipeline.} Vinca has a built-in mechanism to turn a directory of ``package recipes'' into an Azure Build Pipeline definition. Figure~\ref{fig:dependencygraph} shows the dependency graph up to the `ros-core' meta-package. The packages are topographically sorted and the pipeline is then split into multiple stages, where each stage needs to wait for the previous one to finish. Per stage, multiple workers run in parallel with each handling up to five packages. This makes the best use of the ten free Azure workers that run builds in parallel (each worker needs to setup the build environment which takes roughly two minutes -- time that is saved when running multiple builds per worker)\protect\footnote{The number of packages per stage and Azure workers is configurable and will scale if more free workers become available in the future.}. Once the worker is finished, it uploads the packages to the RoboStack channel.

\begin{figure}[t]
  \centering
  \cutpic{0.2cm}{0.98\linewidth}{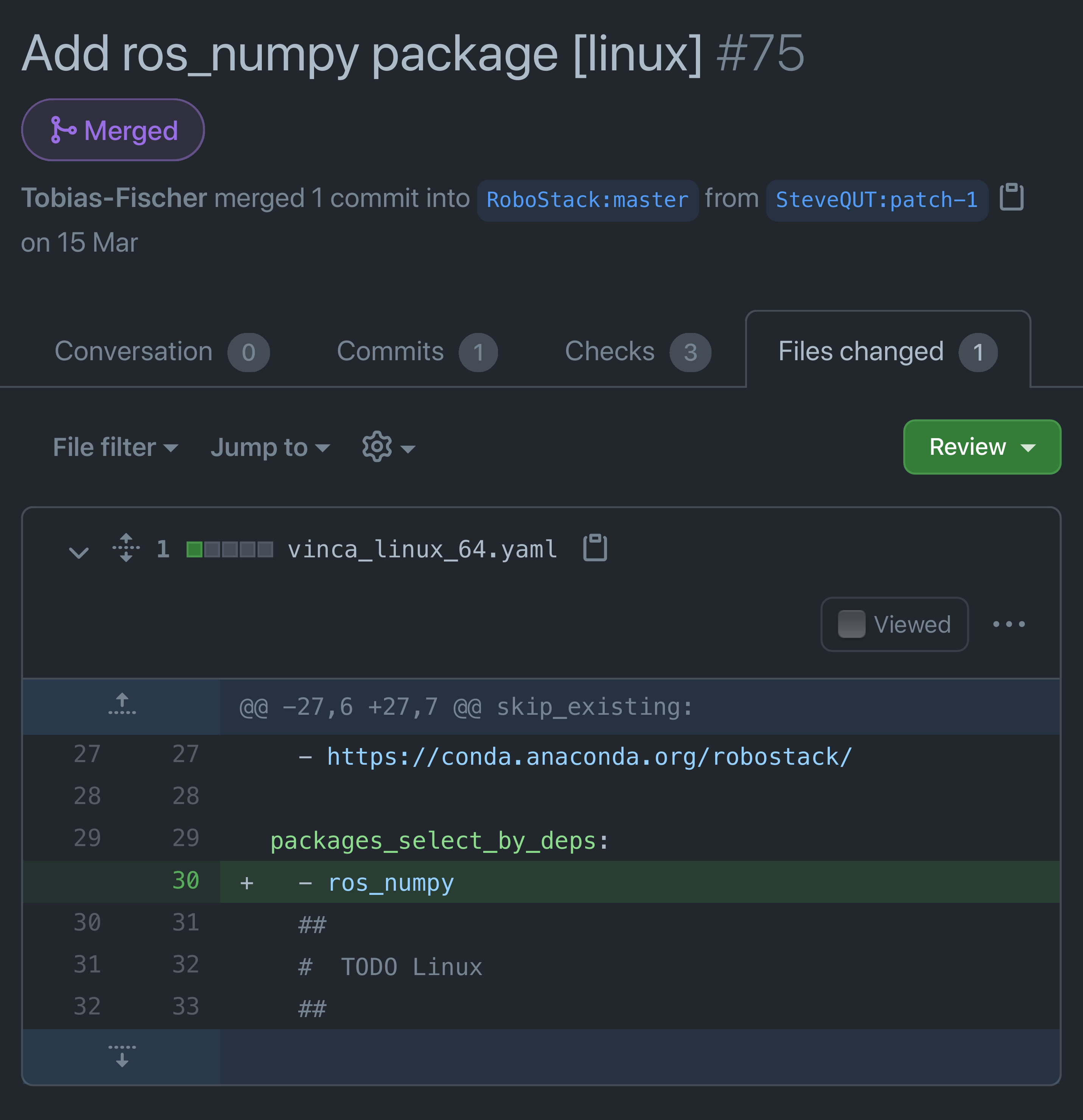}
  \caption{Pull request \#75 which added the ros\_numpy package to the RoboStack channel for Linux by introducing a single new line to the vinca\_linux\_64.yaml configuration file. Other platforms can be targeted in the same way. The pull request is built automatically so that errors can be detected before merging into the main branch.}%
  \label{fig:example_PR}%
\end{figure}



\initialsmall{3}{0.35}{0.3}
\textbf{The third key advance is Mamba, a fully compatible fast drop-in replacement for Conda.} The main motivation for Mamba was to solve the performance issues of Conda, both in terms of speed and memory usage. Conda's dependency resolver is relatively slow, in particular for large software repositories (such as Conda-forge) where the entire repository index is parsed using the standard Python JSON\linebreak
{
\noindent\HorRuleRed
\\[0.2cm]
\color{DarkRed}
\textbf{%
The time taken to solve the dependencies of a new environment containing ros-noetic-desktop-full is reduced to 15s using Mamba, from 90s using Conda.
}
\\
\HorRuleRed
}
parser. Tens of thousands of Python objects are created (as many as there are objects in the JSON index file) and processed before being input in the Boolean satisfiability problem (SAT) solver. On the other hand, Mamba makes use of openSUSE's state-of-the-art libsolv dependency resolver, bypassing most of that processing to resolve dependencies. The core parts of mamba are implemented in C++ for maximum efficiency. Mamba further makes use of multi-threading to download repository data and package files in parallel.

\initialsmall{4}{0.35}{0.3}
\textbf{The fourth key advance is boa, a fast package builder, to build Conda packages.} Boa is re-using a lot of the conda-build infrastructure, but replaces some parts. Specifically, boa uses mamba during the solving stage and has a faster and cleaner recipe-spec implementation.



\subsubsection*{Adding new packages to RoboStack} 
Adding packages that are already released to the ROS build farm can be as simple as adding a single line with the package name to one of the configuration files, as shown in Figure~\ref{fig:example_PR}. The long-term goal is to provide as many packages as possible on all platforms.

\begin{figure}[H]
  \centering
  \includegraphics[width=0.99\linewidth]{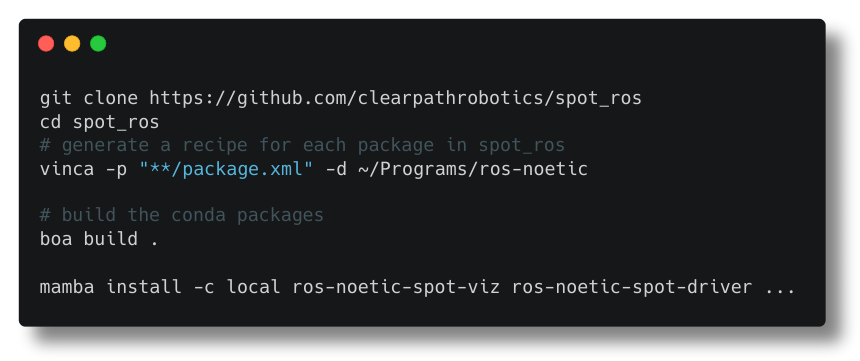}
  \caption{Required commands for cloning the spot\_ros repository, generating recipes for all packages, building and installing them.}%
  \label{fig:custom_packages}%
\end{figure}

RoboStack has the additional advantage that packages can be easily built even if they are not officially released to the ROS servers yet. This mechanism is helpful when the official release is blocked for some reason, but the package itself builds and works fine. Indeed this is important, as~\cite{estefo2019robot} has shown that there is slow process of releasing ROS packages in new ROS distributions. The terminal in Figure~\ref{fig:custom_packages} shows the process for building spot\_ros packages (these packages were at the time of writing not released for ROS Noetic) via the package.xml; this can be easily adapted to other packages.

In a more general use case, where dependencies are not listed in package.xml files, one can use alternative automatic mechanisms (such as ad-hoc Python or CMake scripts) to extract the required metadata and generate Conda binary packages. The authors of this article created an example in the Robotology (not to be confused with RoboStack) CMake-based superbuild~\cite{domenichelli2019build} project that bundles YARP~\cite{Fitzpatrick2014} and other dependencies for the iCub humanoid robot~\cite{metta2010icub}: \url{https://github.com/robotology/robotology-superbuild/pull/652}.

\subsubsection*{Setting up RoboStack in a GitHub Action}
\begin{figure}[H]
  \centering
  \includegraphics[width=0.99\linewidth]{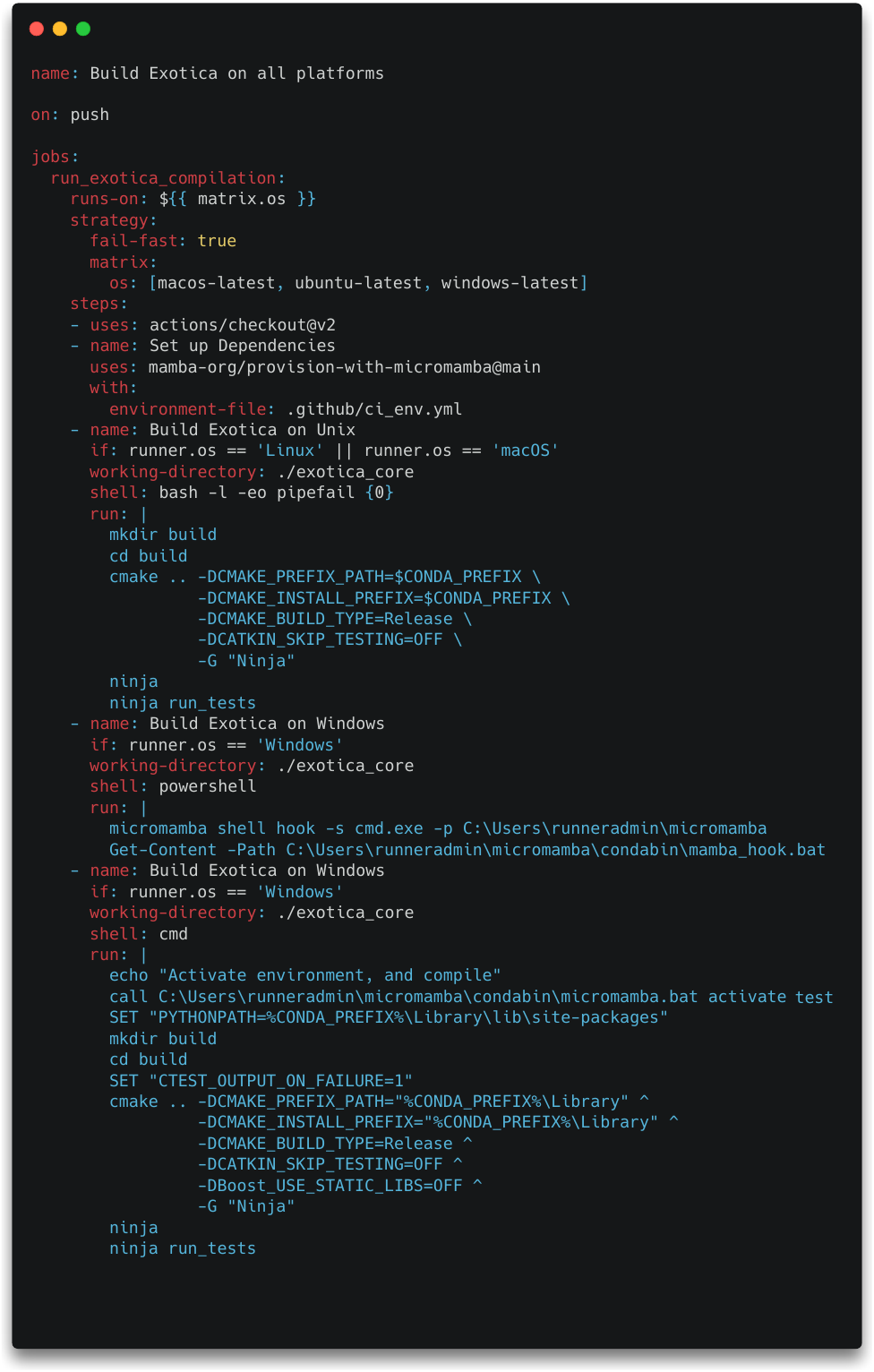}
  \caption{GitHub Action to continuously integrate the exotica\_core package. All dependencies are installed from the Conda-forge and Robostack channels, as specified in the `ci\_env.yml' environment file (not shown for brevity, see Figure~2 of the main tutorial for an example). Adapted from \protect\url{https://github.com/ipab-slmc/exotica/pull/739}}%
  \label{fig:exotica_test_pr}%
\end{figure}

RoboStack can be setup in a GitHub Action for Continuous Integration -- with the benefit that macOS and Windows can be easily tested, which has not been the case for most ROS Continuous Integration pipelines thus far. A full example for the Exotica package that can be straightforwardly adapted is shown in Figure~\ref{fig:exotica_test_pr}.

\newpage
\printbibliography[title={\color{DarkGoldenrod}Appendix references}]